 \documentclass[tablecaption=bottom,wcp]{jmlr} 

\usepackage{algorithm}
\usepackage{algpseudocode}
\usepackage{bbold}
\usepackage{booktabs}
\usepackage{hyperref}
\usepackage{cuted}
\usepackage{physics}

\usepackage{booktabs}
\usepackage[load-configurations=version-1]{siunitx} 

\theorembodyfont{\upshape}
\theoremheaderfont{\scshape}
\theorempostheader{:}
\theoremsep{\newline}

\jmlrproceedings{}{}

\title[Penalty Bayesian Neural Networks]{Data Subsampling for Bayesian Neural Networks}

\author{\Name{Eiji Kawasaki}\and\Name{Lawrence Adu-Gyamfi}\\
 \addr Université Paris-Saclay, CEA, List, F-91120, Palaiseau, France
 \AND
 \Name{Markus Holzmann}\\
 \addr Univ. Grenoble Alpes, CNRS, LPMMC, 38000 Grenoble, France
}

\begin{document}

\maketitle

\begin{abstract}
Markov Chain Monte Carlo (MCMC) algorithms do not scale well for large datasets leading to difficulties in Neural Network posterior sampling. In this paper, we propose Penalty Bayesian Neural Networks - PBNNs, as a new algorithm that allows the evaluation of the likelihood using subsampled batch data (mini-batches) in a Bayesian inference context towards addressing scalability. PBNN avoids the biases inherent in other naive subsampling techniques by incorporating a penalty term as part of a generalization of the Metropolis Hastings algorithm. We show that it is straightforward to integrate PBNN with existing MCMC frameworks, as the variance of the loss function merely reduces the acceptance probability. By comparing with alternative sampling strategies on both synthetic data and the MNIST dataset, we demonstrate that PBNN achieves good predictive performance even for small mini-batch sizes of data. We show that PBNN provides a novel approach for calibrating the predictive distribution by varying the mini-batch size, significantly reducing predictive overconfidence.
\end{abstract}

\section{INTRODUCTION}

The development of robust and reliable Deep Neural Networks (DNNs) has led to a significant interest in the application of Uncertainty Quantification (UQ) techniques. UQ is essential for assessing the confidence in predictions made by DNNs, especially in critical applications such as autonomous driving and medical diagnosis.
Despite its importance, designing effective UQ methods that accurately compute the predictive distribution by marginalizing over DNN parameters remains a challenge \citep{gawlikowski_survey_2023}.

Bayesian inference methods provide a principled approach to obtain the posterior of network parameters. These methods, which include Variational inference and Markov Chain Monte Carlo (MCMC) sampling, the latter regarded as being the gold standard \citep{pmlr-v139-izmailov21a}, face significant scalability issues. Specifically, MCMC’s requirement to evaluate the log-likelihood over the entire dataset at each iteration makes it impractical for large systems and datasets. This challenge hinders the practical application of such posterior sampling techniques in Bayesian Neural Networks (BNNs). Additionally, uninformative priors are commonly used to prevent overconfidence (overfitting), adversely affecting the predictive performance of BNNs.  A current field of research is dedicated to developing BNN-specific priors as reviewed by \citet{https://doi.org/10.1111/insr.12502}.

Towards addressing these challenges, we propose a variant of MCMC which we call Penalty Bayesian Neural Network (PBNN). \textbf{PBNN leverages a novel approach of using subsampled batch data (mini-batches) while incorporating a "penalty method" for BNN posterior sampling. This strategy significantly reduces predictive overconfidence and avoids the biases inherent in naive subsampling techniques.} PBNN adapts the penalty method \citep{10.1063/1.478034}, originally developed in the context of statistical and computational physics, e.g. in the work of \citet{PhysRevLett.93.146402}, to efficiently sample distributions with loss functions affected by statistical noise. The use of this method in PBNN enables unbiased posterior sampling by explicitly accounting for the variance of the loss associated with subsampled data batches (mini-batches).

The issue of introduction of strong bias in the posterior sampling process when done naively has been explored in previous work in the context of Bayesian inference resulting in the proposal of several methodologies addressing scalability  \citep{bardenet_markov_2017,pmlr-v32-korattikara14,ijcai2018p753}. PBNN distinguishes itself by ensuring unbiased sampling of the posterior for any given mini-batch size under the assumption that the
distribution generated by the mini-batches is normal and of known
variance. Although the later condition might be slightly violated in practice, as the variance  must be
estimated itself, our approach represents a departure from other methods such as SGMCMC (HMC and Langevin) and Barker acceptance test that attempt to control bias post-sampling and reducing it below a threshold.

The remainder of the article is structured as follows: we begin with an overview of fundamental concepts and intuitions underlying the sampling of posterior distributions through Bayesian inference, we then delve into some related works, including a review of the Stochastic Gradient Langevin Dynamic (SGLD) algorithm \citep{10.5555/3104482.3104568} and its implications for noise in loss computation. Subsequently, we introduce PBNN as a novel, unbiased posterior sampling strategy and show its benefits focusing on its ability to effectively mitigate the predictive overconfidence effects.

We proceed to discuss some practical details on evaluating the noisy loss, and its uncertainty, highlighting the compatibility of PBNN with state of the art MCMC proposal distribution for the Markov chain such as Langevin samplers.
We demonstrate PBNNs predictive performance by testing it on a learning task designed to trigger overfitting. This demonstration not only highlights PBNN’s effectiveness but also introduces the mini-batch size as an efficient calibration parameter.
We conclude by discussing the impact of mini-batch size, as a parameter, on acceptance rates and overall performance.

\section{BACKGROUND}
\label{section: PRELIMINARIES}

In the following, we consider a vector $\theta$ that describes the parameters of a model, representing the weights and biases of a neural network hereafter. We define $p(\theta)$ as a prior distribution over this set of parameters. Commonly used priors are Gaussian and Laplace that correspond respectively to L2 and L1 regularizations. We refer to $p(y|x,\theta)$ as the probability of a target $y$ given a data input $x$ and a parameter vector $\theta$.
The uncertainty over the parameters $\theta$ given a collection of observed data $\mathcal{D}$ is captured by the posterior distribution which is defined as:
\begin{equation}
    p(\theta|\mathcal{D})=\frac{p(\theta)\prod_{i=1}^N p(y_i|x_i,\theta)}{p(\mathcal{D})}
    \label{equation: posterior distribution}
\end{equation}
where the observations in the data set $\mathcal{D}=\left\{(y_i, x_i)\right\}_{i=1}^N$ are iid. Up to a constant, $p(\theta|\mathcal{D})\propto e^{-\mathcal{L}_N^N(\theta)}$ where the loss $\mathcal{L}_N^N(\theta)$ is the negative log of the posterior:
\begin{equation}
    \mathcal{L}_N^N(\theta) = -\log p(\theta)-\frac{N}{N}\sum_{i=1}^N\log p(y_i|x_i,\theta)
    \label{equation: BNN loss}
\end{equation}
and the last term is the negative log-likelihood. This choice of the loss is only illustrative and does not reduce the generality of PBNN as we could have also considered an unsupervised setup where $\mathcal{D}=\left\{(x_i)\right\}_{i=1}^N$ and $\mathcal{L}_N^N(\theta) = -\log p(\theta)-\sum_{i=1}^N \log p(x_i|\theta)$.

As an example of a common loss, using a one-dimensional homoscedastic Gaussian likelihood in a regression task and a Gaussian prior over the model's parameters, we obtain the familiar squared-error loss function:
\begin{equation}
    \mathcal{L}_N^N(\theta)\propto\lambda\lVert\theta\rVert^2+\sum_{i=1}^N(y_i-f_\theta(x))^2+\text{cst}
    \label{equation: MSE}
\end{equation}
where $f_\theta(x)$ is the prediction of the neural network and $\lambda$ tunes the strength of the L2 regularization.

The parameters $\theta$ of the network are usually optimized through Stochastic Gradient Descent (SGD), as defined below: 
\begin{equation}
    \theta_{t+1}= \theta_t -\eta\nabla_{\theta} \mathcal{L}_n^N(\theta_t)
    \label{equation: SGD}
\end{equation}
$\eta\in\mathbb{R}^+$ is called the learning rate. Stochastic gradient descent algorithms aim at maximizing the posterior distribution (MAP) using mini-batches of data from the whole dataset $\mathcal{D}$. The new stochastic loss $\mathcal{L}_n^N(\theta_t)$ is defined as:
\begin{equation}
    \mathcal{L}_n^N(\theta)=-\log p(\theta)-\frac{N}{n}\sum_{i=1}^n \log p(y_i|x_i,\theta)
    \label{equation: mini-batch loss}
\end{equation}
where $n\leq N$ corresponds to the size of the mini-batch. This mini-batch contains data sampled without replacement from dataset $\mathcal{D}$ such that by definition $\langle\mathcal{L}_n^N(\theta)\rangle=\mathcal{L}_N^N(\theta)$.

In order to capture the epistemic uncertainty of $\theta$ given $\mathcal{D}$, we want to obtain samples $\theta$ that are distributed according to \equationref{equation: posterior distribution} instead of a point estimate $\theta_{t\rightarrow\infty}$ that minimizes the loss. In the special case of deep learning, the class of models and techniques that target this posterior distribution are called Bayesian Neural Networks (BNNs).

As standard in Bayesian inference, the prediction of a BNN corresponds to a distribution of possible values $y$ given an input $x$ and conditional on the training data set $\mathcal{D}$, defined as:
\begin{equation}
    p(y|x,\mathcal{D})= \int p(y|x, \theta)p(\theta|\mathcal{D})d\theta
    \label{equation: predictive posterior}
\end{equation}
This is called the predictive posterior distribution and is calculated by marginalizing the product of the likelihood and posterior distribution over the model's parameters. We can approximate this distribution by computing a Monte Carlo estimate of an expected value:
\begin{equation}
    p(y|x,\mathcal{D})=\mathbb{E}_{\theta\sim p(\theta|\mathcal{D})}[p(y|x, \theta)]\simeq\frac{1}{L}\sum_{i=1}^L p(y|x, \theta^{(i)})        
    \label{equation: predictive distribution}
\end{equation}
where $\theta^{(1)}, ..., \theta^{(L)}$ are $L$ independent and identically distributed (iid) random parameters sampled from the posterior distribution that we can obtain for instance using a MCMC algorithm. 

In order to calibrate the predictive posterior, one can use Safe Bayes approaches \citep{wilson_bayesian_2020} such as considering $n\in\mathbb{R}^+$ as a tunable hyper-parameter without using any data subsampling strategy. The loss then writes:
\begin{equation}
    \mathcal{L}_N^n(\theta)=-\log p(\theta)-\frac{n}{N}\sum_{i=1}^N \log p(y_i|x_i,\theta)
    \label{equation: mini-batch loss}
\end{equation}
This technique is called tempering, as the $N/n$ factor can be interpreted as a temperature using the analogy with statistical physics. The temperature parameter is known to help in cases of model \textbf{misspecification} where there is no $\theta$ for which it can be assumed that the observations $(y_i,x_i)$ are iid according to $p(y|x,\theta)$.

\section{RELATED WORK}

In the context of sampling the posterior distribution, we introduce in this section some relevant literature that studied how to take into account a noisy estimate of either $\mathcal{L}_N^N(\theta)$ or $\nabla_{\theta}\mathcal{L}_N^N(\theta)$ computed from a subset of the data. The link between neural network noisy gradient and sampling a BNN posterior distribution is discussed in \sectionref{section: Penalized Langevin Dynamic}.

\paragraph{Stochastic Gradient Langevin Dynamics}
\citet{10.5555/3104482.3104568} showed that the iterates $\theta_t$ will converge to samples from the target posterior distribution in \equationref{equation: posterior distribution} as they anneal the stepsize by adding the right amount of noise to a standard stochastic gradient optimization algorithm. This is known as the Stochastic Gradient Langevin Dynamics (SGLD) where the parameter update is given by:
\begin{equation}
    \theta_{t+1}= \theta_t -\eta_t\nabla_{\theta} \mathcal{L}_n^N(\theta_t) + \sqrt{2\eta_t}\epsilon_t
    \label{equation: SGLD}
\end{equation}
where $\epsilon_t$ is a random vector containing centered normally distributed components. No rejection step is required for a vanishing step size. \citet{pmlr-v32-cheni14} later extended this idea to HMC sampler.

\paragraph{Noisy Posterior Sampling Bias} Due to a potentially high variance of the stochastic gradients $\nabla_{\theta}\mathcal{L}_n^N(\theta_t)$, \citet{brosse_promises_2018} showed that the SGLD algorithm has an invariant probability measure which in general significantly departs from the target posterior for any non vanishing stepsize $\eta_t$. Furthermore, a recent work from \citet{DBLP:journals/corr/abs-2102-01691} suggests that recent versions of SGLD  implementing an additional Metropolis-Hastings rejection step do not improve this issue, because the resulting acceptance probability is likely to vanish too.

\paragraph{Adaptive Subsampling Approach}\citet{pmlr-v32-bardenet14} showed one way to cope with the bias that is introduced by the data subsampling in the MCMC computation of a BNN posterior distribution. They propose an approximate implementation of the accept/reject step of Metropolis-Hastings algorithm while providing guarantees to coincide with this step based on the full dataset with a probability superior to a user-specified tolerance level.

\paragraph{Failures of Data Set Splitting Inference} 
Other works have exploited parallel computing to scale Bayesian inference to large datasets by using a two-step approach. First, a MCMC computation is run in parallel on $K$ (sub)posteriors defined on data partitions following $p(\theta|\mathcal{D})\propto\prod_{i=1}^K p(\theta)^{1/K}e^{-\mathcal{L}_{n,i}^{n}}$, where $i$ corresponds to the index of a randomly chosen mini-batch containing $n=N/K$ data. Then, a server combines local results. While efficient, this framework is very sensitive to the quality of sub-posterior sampling as shown by \citet{souza_parallel_2022}.

\section{PENALTY BAYESIAN NEURAL NETWORK}

\subsection{Biased Posterior Sampling Because Of Mini-Batches}
\label{subsection: Unbiased posterior sampling}

We can draw iid samples from the posterior distribution defined in \equationref{equation: posterior distribution} using MCMC through exploration of the state space of $\theta$ using Markov chains.
It is well known that the detailed balance equation is a sufficient but not necessary condition ensuring that this Markov process possesses a stationary distribution corresponding to \equationref{equation: posterior distribution}. Detailed balance is given by:
\begin{equation}
    A(\theta,\theta')q(\theta|\theta')e^{-\Delta({\theta',\theta})}=A(\theta',\theta)q(\theta'|\theta)
    \label{equation: original detailed balance}
\end{equation}
where $A(\theta', \theta)$ corresponds to the probability of accepting a move from the parameter set $\theta$ to $\theta'$ suggested by the proposal distribution $q(\theta'|\theta)$. For the sake of brevity in notation we consider in this section a random walk algorithm corresponding to drawing a centered reduced normal random variable $\epsilon$ and computing $\theta'=\theta+\sqrt{2\eta}\epsilon$ such that the proposal distribution is symmetric $q(\theta'|\theta)=q(\theta|\theta')$. Using the Metropolis–Hastings (MH) algorithm, the acceptance then writes:
\begin{equation}
    A(\theta',\theta)=\min\left(1,e^{-\Delta({\theta',\theta})}\right)\qquad\text{with}\ \Delta({\theta',\theta}) = \mathcal{L}_N^N(\theta') - \mathcal{L}_N^N(\theta)
    \label{equation: MH usual acceptance}
\end{equation}

We would rather compute the loss differences over random mini-batches instead of the full dataset $\mathcal{D}$. Consequently, we introduce a random variable $\delta({\theta',\theta})$:
\begin{equation}
    \delta({\theta',\theta}) = \mathcal{L}_n^N(\theta') - \mathcal{L}_n^N(\theta)
    \label{equation: usual delta}
\end{equation}
that is a noisy estimate of the loss defined in \equationref{equation: MH usual acceptance} such that $\Delta({\theta',\theta})=\langle\delta({\theta',\theta})\rangle$. As shown in the \figureref{figure: penalty_linear_regression,fig:without_penalty_batch_size_10}, naively replacing $\Delta$ by $\delta$ in \equationref{equation: MH usual acceptance} prevents MCMC algorithms from accurately sampling the target posterior distribution. We need to take into account the noise in the updated loss difference $\delta({\theta',\theta})$ caused by subsampling the data. This is in stark contrast to SGD algorithms (cf. \equationref{equation: SGD}) where a similar stochasticity has been shown to be beneficial for the optimization.

\subsection{Noise Penalty Theory}

In the context of statistical physics and computational chemistry, \citet{10.1063/1.478034} generalized the Metropolis-Hastings random walk algorithm to the situation where the energy difference $\Delta$ (loss difference in our case) is noisy and can only be estimated. They showed that it is possible to sample the target distribution if we assume that we can write $\delta({\theta',\theta})$ as equal to the loss difference $\Delta({\theta',\theta})$ plus a Gaussian noise, meaning:
\begin{equation}
    \delta({\theta',\theta})\sim\mathcal{N}(\delta;\Delta({\theta',\theta}),\sigma^2({\theta',\theta}))
    \label{equation: formal definition}
\end{equation}
We can sample the target distribution under significant Gaussian noise by applying a penalty term $-\sigma^2({\theta',\theta})/2$ to the noisy difference in the acceptance ratio $A$ such that
\begin{equation}
    A(\theta',\theta)=\min\left(1,e^{-\delta({\theta',\theta})-\sigma^2({\theta',\theta})/2}\right)
    \label{equation: random walk penalty acceptance}
\end{equation}
One can then show that detailed balance is satisfied on average, which is a sufficient condition for the Markov chain
to sample the target distribution.
\begin{equation}
    \int d\delta A(\theta,\theta')\mathcal{N}(\delta;\Delta({\theta',\theta}),\sigma^2({\theta',\theta}))=e^{-\Delta(\theta',\theta)}\int d\delta A(\theta',\theta)\mathcal{N}(\delta;\Delta({\theta,\theta'}),\sigma^2({\theta,\theta'}))
    \label{equation: noisy detailed balance}
\end{equation}

In order to demonstrate the validity of the noise penalty method, let us first consider the case where the
integration over the noise $\delta$ is performed exactly, either analytically or via numerical quadrature. Then, a Monte Carlo proposition $\theta \to \theta'$ is accepted with probability
\begin{equation}
\overline{A}(\theta',\theta) = \int d \delta  
\mathcal{N}(\delta;\Delta({\theta',\theta}),\sigma^2({\theta',\theta}))  A(\theta',\theta)
\label{equation: noisy averaged A}
\end{equation}
and a Markov chain Monte Carlo based on $\overline{A}(\theta',\theta)$ satisfies
the detailed balance condition, Eq.~(\ref{equation: noisy detailed balance}), or
\begin{equation}
\overline{A}(\theta',\theta) p(\theta|\mathcal{D}) = \overline{A}(\theta,\theta') p(\theta'|\mathcal{D})
\end{equation}
as can be checked explicitly \cite{10.1063/1.478034}. Under the usual assumption of irreducibility
of $\overline{A}(\theta',\theta)$, the corresponding MCMC converges to a unique stationary 
distribution, $p(\theta|\mathcal{D})$. However, this result of the MCMC cannot depend on the
underlying quadrature used to perform the integration over the noise, and a statistical
evaluation of Eq.~(\ref{equation: noisy averaged A}) at each step of the Markov chain
must converge to the same distribution.
Therefore, the noise penalty methods gives access to an unbiased sampling of $p(\theta|\mathcal{D})$ under our assumption of a normal distribution of $\delta(\theta',\theta)$ with known variance $\sigma^2(\theta',\theta)$.

\begin{figure}[!ht]
    \centering
    \includegraphics[width=.5\textwidth]{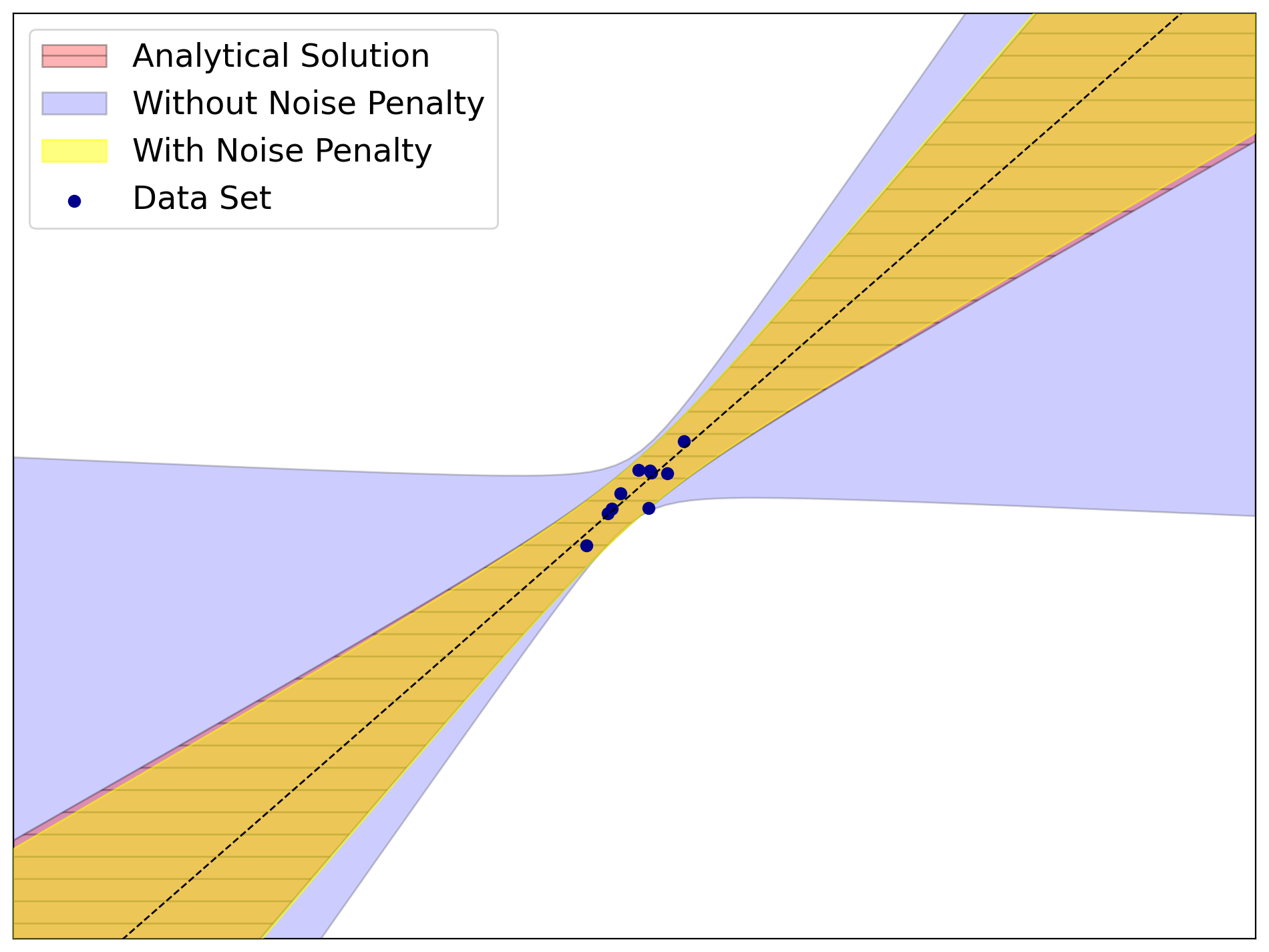}
    \caption{Illustration of posterior predictive distributions, defined in \equationref{equation: predictive distribution}, computed for a linear regression task. The coloured areas correspond to the mean of the distributions $\pm$ one standard deviation. The blue area is computed by naively replacing $\Delta$ by $\delta$ in the MH algorithm from \equationref{equation: MH usual acceptance}. The noisy loss difference $\delta$ as defined by \equationref{equation: usual delta} is computed on a single mini-batch of 2 data points. The yellow area shows a similar computation, that incorporates the noise penalty term as defined in \equationref{equation: random walk penalty acceptance}. It correctly matches the red area corresponding to the analytical Bayesian linear regression with Gaussian prior and known variance \citep{bishop2007}.}
    \label{figure: penalty_linear_regression}
\end{figure}

\figureref{figure: penalty_linear_regression} shows the benefits of adding the noise penalty on a synthetic linear regression example. Typically, increasing the size $n$ of the mini-batches diminishes the magnitude of the noise, i.e. reduces the variance $\sigma^2({\theta',\theta})$ and therefore increases the acceptance $A(\theta',\theta)$.

From \equationref{equation: random walk penalty acceptance} one can immediately identify a drawback of PBNN leading to an exponential reduction of the acceptance as the variance $\sigma^2({\theta',\theta})$ is always non-negative. In \sectionref{section: Penalized Langevin Dynamic}, we show an extension of the penalty method to non-symmetric proposal distributions $q(\theta'|\theta)$ \citep{10.5555/3104482.3104568, neal_mcmc_2010} that aims at increasing the acceptance, thus mitigating this drawback.
Additionally, in the case of BNNs, the variance $\sigma^2({\theta',\theta})$ is unknown and can only be estimated. While it is possible to account for noisy variances \citep{10.1063/1.478034}, we will not explore this extension further.

\subsection{Expected Posterior Sampling Using Mini-Batches}
\label{subsection: Expected posterior sampling with mini-batches}

As shown in \figureref{figure: penalty_linear_regression}, it is possible to target the usual posterior distribution as defined by the loss given in \equationref{equation: BNN loss} using the noise penalty. Since the noise penalty requires the computation of $\sigma^2({\theta,\theta'})$, i.e. the expected variance of the loss differences over multiple mini-batches, we would like to show in the following that it is more interesting to target the posterior distribution defined by the mean loss $\mathcal{L}_n(\theta)$. This loss corresponds to the average loss given an infinite number of mini-batches.
\begin{equation}
    \mathcal{L}_n(\theta)=\langle\mathcal{L}_n^n(\theta)\rangle
\end{equation}
One can write this mean loss using the expected value of the likelihood as shown in \equationref{equation: expected loss}. In practice, computing this quantity is intractable since it would require an infinite amount of data.
\begin{equation}
    \mathcal{L}_n(\theta) = -\log p(\theta)-n\mathbb{E}[\log p(y_i|x_i,\theta)]
    \label{equation: expected loss}
\end{equation}

It is important to distinguish between $\mathcal{L}_n(\theta)$ and $\mathcal{L}_N^N(\theta)=\langle\mathcal{L}_n^N(\theta)\rangle$, where the latter represents the mean loss over all mini-batches within a single dataset. Specifically, when $n=N$,  computing $\mathcal{L}_N(\theta)$ is comparative to treating the entire training dataset as a single large mini-batch and computing its expected loss as $\langle\mathcal{L}_{N}^{N}(\theta)\rangle=\mathcal{L}_N(\theta)$. Applying the MH algorithm as defined by \equationref{equation: MH usual acceptance}  is impossible as the loss difference $\Delta$ given by:
\begin{equation}
    \Delta({\theta',\theta}) = \mathcal{L}_n(\theta') - \mathcal{L}_n(\theta)
\end{equation} 
is unknown and generally intractable.
We can, however, compute a noisy estimate $\delta$ of $\Delta$ based on the losses of the mini-batches:
\begin{equation}
    \delta({\theta',\theta})=\mathcal{L}_{n}^n(\theta')-\mathcal{L}_{n}^n(\theta)
    \label{equation: mini-batch loss difference}
\end{equation}
where it is evident that $\langle\delta({\theta',\theta})\rangle=\Delta({\theta',\theta})$. In this case, varying the size $n$ of the mini-batches corresponds to changing the target poster distribution that we wish to sample. This relates to empirical Bayes and tempering techniques shown in the appendix (see \sectionref{apd:first}). However, these well establish techniques do not target the distribution defined by \equationref{equation: expected loss}, resulting in qualitatively different results compared to PBNN (cf. \figureref{figure: tempered} in \appendixref{apd:first}). As shown in \equationref{equation: mini-batch loss}, tempering the posterior distribution requires the computation of the likelihood over the full data set, a process we aim to avoid.

\begin{figure}[!ht]
    \centering
    \subfigure[BNN prediction corresponding to\newline 
    \equationref{equation: predictive posterior} with $N=2000$.]{
        \includegraphics[width=.45\textwidth]{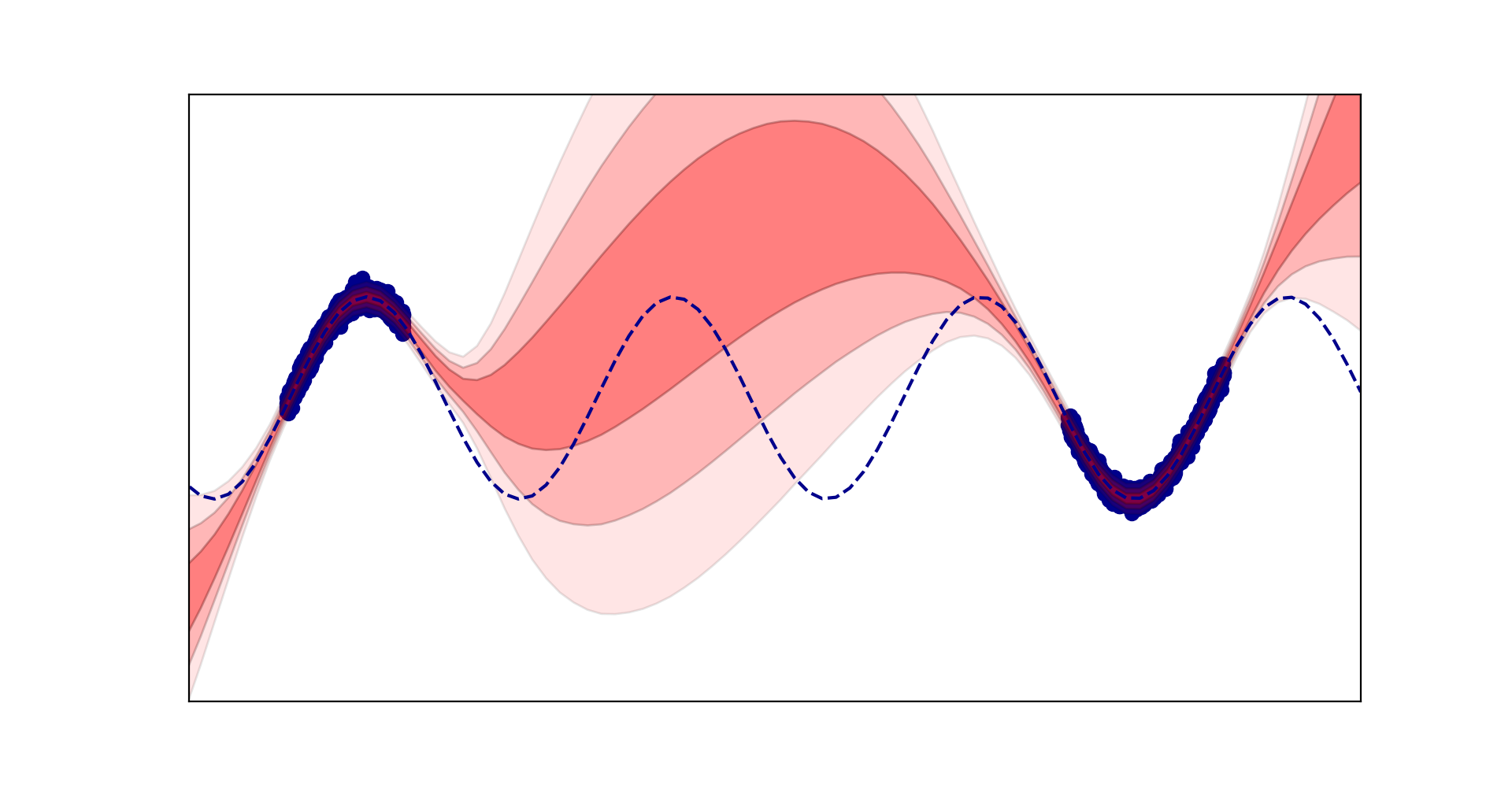}
        \label{fig:bnn_regression}
    }
    \subfigure[PBNN prediction,\ $n=5$ and $M=5$.]{
        \includegraphics[width=.45\textwidth]{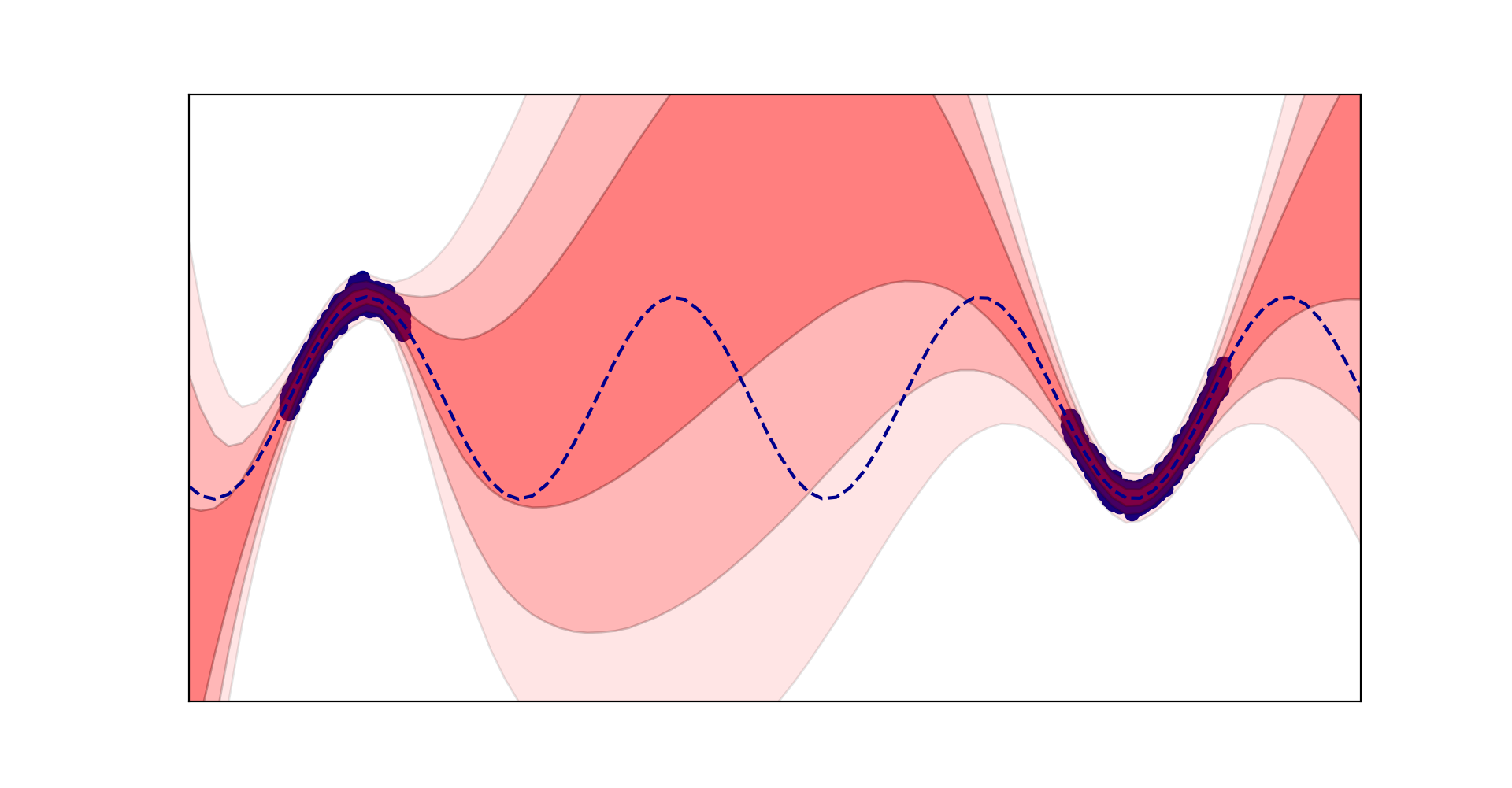}
        \label{fig:penalty_regression_batch_size_5}
    }
    \subfigure[Prediction without noise penalty\newline using \equationref{equation: usual delta}, $n=10$.]{
        \includegraphics[width=.45\textwidth]{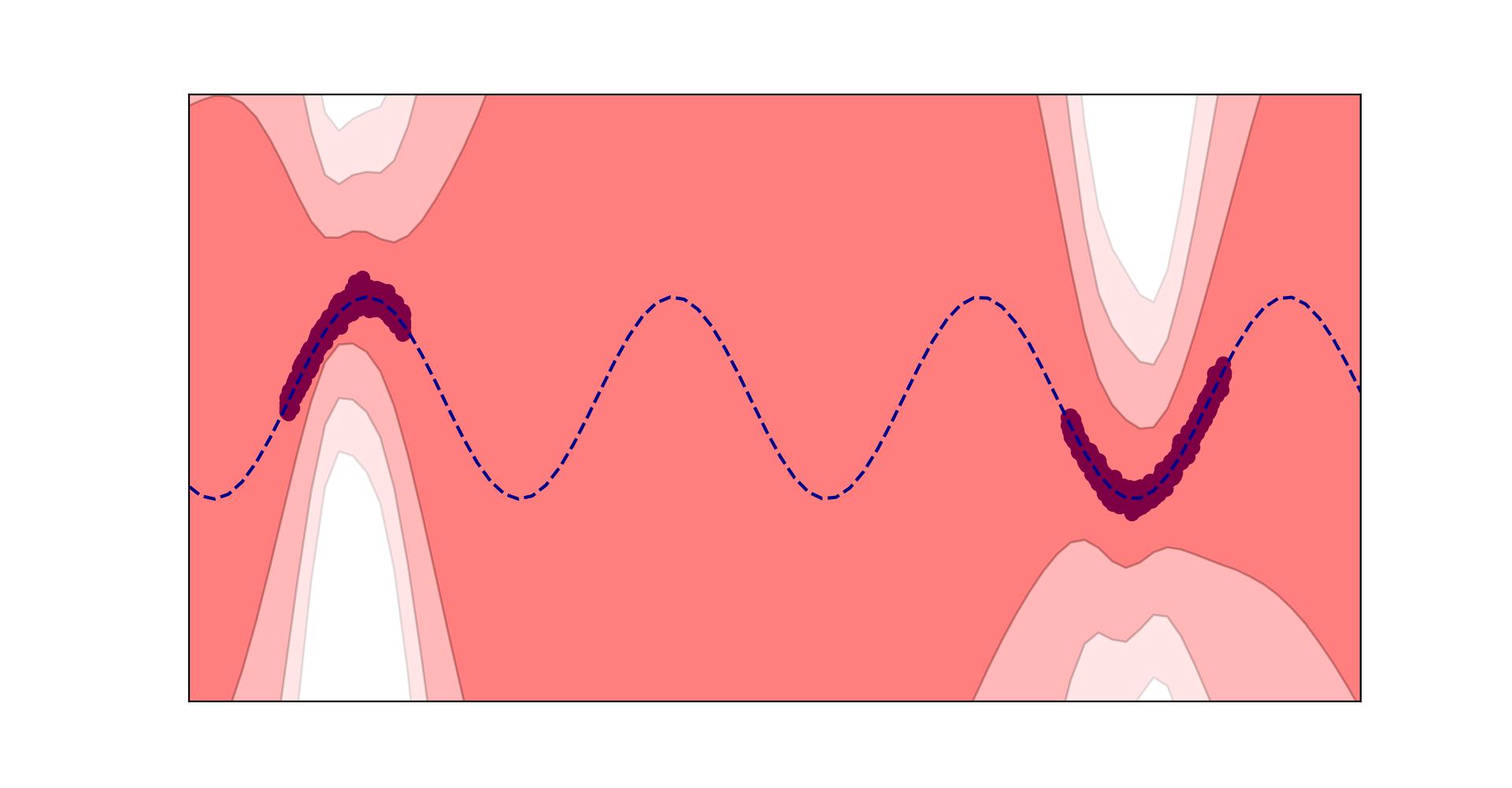}
        \label{fig:without_penalty_batch_size_10}
    }
    \subfigure[PBNN prediction,\ $n=10$ and $M=5$.]{
        \includegraphics[width=.45\textwidth]{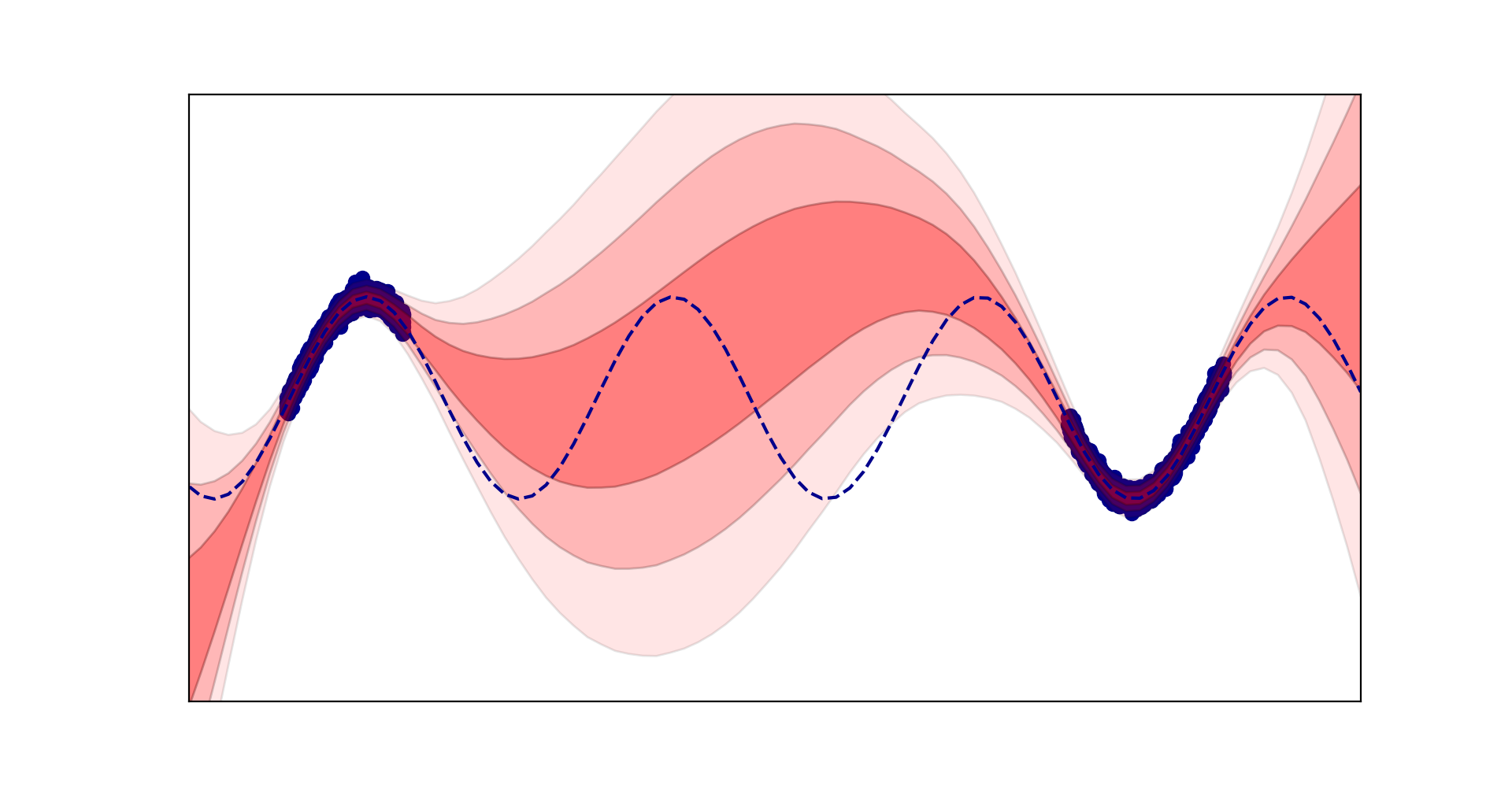}
        \label{fig:penalty_regression_batch_size_10}
    }
    \caption{Illustration of posterior predictive distributions, defined in \equationref{equation: PBNN predictive posterior}. Shaded regions indicate predictive means ± one, two and three standard deviations. We use a homoscedastic Gaussian likelihood (cf. loss defined in \equationref{equation: MSE}). The noise in the data is small such that the visible posterior distribution variance is due to the epistemic uncertainty. The training data set contains $N=2000$ points meaning that at each step, only a fraction of the data points is used by the PBNN. We emphasize that none of these images are expected to match \figureref{fig:bnn_regression} as each targets a different posterior distribution.}
    \label{figure: synthetic_regression}
\end{figure}

\textbf{The noise penalty method is particularly relevant for BNNs as these networks do not scale well for large dataset. Additionally, their common uninformative prior makes it difficult to calibrate the Bayesian predictive marginal distribution.} Similar to Safe Bayes approaches (see \sectionref{apd:first}), one can use the mini-batch size $n$ as a calibration hyper-parameter. The example on a synthetic non-linear regression task from \figureref{figure: synthetic_regression} shows the benefits of targeting the posterior distribution given by \equationref{equation: expected loss} instead of \equationref{equation: BNN loss}. It shows an example of a MCMC algorithm that uses a noise penalty in the acceptance \equationref{equation: PBNN MH acceptance}. The figure shows the expected posterior predictive distributions defined as:
 \begin{equation}
    p(y|x,n)= \dfrac{\displaystyle\int p(y|x, \theta)e^{-\mathcal{L}_n(\theta)}d\theta}{\displaystyle\int e^{-\mathcal{L}_n(\theta)}d\theta}\simeq\frac{1}{L}\sum_{i=1}^L p(y|x, \theta^{(i)})
    \label{equation: PBNN predictive posterior}
\end{equation}
where $\theta^{(i)}$ should be iid samples obtained from the posterior distribution which is proportional to $e^{-\mathcal{L}_n(\theta)}$. In practice, obtaining such a set of samples using a MCMC algorithm as defined in \equationref{equation: PBNN MH acceptance} challenges the iid assumption since $\delta({\theta',\theta})$ are computed on a limited amount of mini-batches. As shown in the rest of this article, empirical results show that this assumption is not an obstacle to the use of this method.

Even in the situation where $\delta({\theta',\theta})$ is not distributed according to a Gaussian distribution as required by \equationref{equation: formal definition}, we can estimate its variance following:
\begin{equation}
    \sigma^2({\theta',\theta})\simeq\frac{1}{M-1}\sum_{i=1}^M\left(\mathcal{L}_{n,i}^n(\theta')-\mathcal{L}_{n,i}^n(\theta)-\delta({\theta',\theta})\right)^2
    \label{equation: mini-batch loss difference variance}
\end{equation}
where $i$ denotes the index of a randomly chosen mini-batch.

\subsection{Large Number Of Mini-Batches $M$ Scenario}
\label{subsection: Noise penalty estimation}

In the case of a large number of mini-batches $M$ i.e a large available dataset and small mini-batches $n$, it is convenient to define $\delta({\theta',\theta})$ as an empirical average over the loss difference:
\begin{equation}
    \delta({\theta',\theta})=\frac{1}{M}\sum_{i=1}^M \left(\mathcal{L}_{n,i}^n(\theta')-\mathcal{L}_{n, i}^n(\theta)\right)
    \label{equation: empirical average over the loss difference}
\end{equation}
with $M$ strictly smaller than the total number of available mini-batches $N/n$. By definition, this average is an unbiased estimator such that $\langle\delta({\theta',\theta})\rangle=\Delta({\theta',\theta})$. Importantly, the central limit theorem ensures that $\delta({\theta',\theta})$ approaches a normal distribution for large M as required by the noise penalty setup in the \equationref{equation: formal definition}.

Using the definition from \equationref{equation: empirical average over the loss difference}, the variance of the random variable $\delta({\theta',\theta})$ strictly decreases with the number of mini-batches $M$. This is very convenient since, as shown in \equationref{equation: random walk penalty acceptance}, the loss difference $\delta({\theta',\theta})$ must dominate the always positive variance $\sigma^2({\theta',\theta})/2$ in order to obtain a reasonable acceptance $A(\theta', \theta)$ i.e sufficiently greater than zero.

While $\sigma^2({\theta',\theta})$ is unknown, we can obtain an unbiased estimate for it as:
\begin{equation}
    \sigma^2({\theta',\theta})\simeq\frac{1}{M(M-1)}\sum_{i=1}^M\left(\mathcal{L}_{n,i}^n(\theta')-\mathcal{L}_{n,i}^n(\theta)-\delta({\theta',\theta})\right)^2
    \label{equation: chi-squared}
\end{equation}
In the following, we do not take into account the error over the estimation of the variance $\sigma^2({\theta',\theta})$. This assumption is based on the hypothesis that the variations of $\sigma^2$ as a function of $\theta'$ and $\theta$ predominate over the estimation noise. Alternatively, a significant number of mini-batches is required to minimize this error. Leading order corrections in this noise are discussed in \citet{10.1063/1.478034}.

\section{PBNN ALGORITHM}

\subsection{Penalty Bayesian Neural Network Posterior Sampling Algorithm}

Given \equationref{equation: empirical average over the loss difference} and \equationref{equation: chi-squared} we can design \algorithmref{algorithm: Penalty Method Algorithm} that is able to sample BNN posterior distributions while restricting the evaluation of the likelihood to $n$ data points and $M$ small mini-batches at each iteration step. \figureref{fig:penalty_regression_batch_size_5} and \figureref{fig:penalty_regression_batch_size_10} are drawn using this algorithm.

\begin{algorithm}[!ht]
\caption{PBNN Algorithm}
    \vspace{.2cm}
    \begin{enumerate}
        \item Initialize a parameter vector $\theta_t \gets \theta_0$
        \item For $t=1$ to maximum number of iterations
        \begin{enumerate}
            \item \label{step: sample proposal}Sample a new configuration $\theta' \sim q(\theta'|\theta_t)$
            \item Compute the noisy loss difference $\delta({\theta',\theta_t})$ from \equationref{equation: empirical average over the loss difference} (or \equationref{equation: mini-batch loss difference})
            \item Compute the corresponding $\sigma^2({\theta',\theta_t})$ from \equationref{equation: chi-squared} (or \equationref{equation: mini-batch loss difference variance})
            \item Compute the acceptance $\displaystyle A(\theta', \theta_t) \gets \min\left( 1,\frac{q(\theta_t|\theta')}{q(\theta'|\theta_t)}e^{-\delta({\theta',\theta_t})-\sigma^2({\theta',\theta_t})/2}\right)$
            \item Sample a uniform random variable $u \sim \mathcal{U}(0,1)$
            \item If $u\leq A(\theta', \theta_t)$
            \begin{enumerate*}
                \item[] $\theta_{t+1}\gets\theta'$
            \end{enumerate*}
            otherwise
            \begin{enumerate*}
                \item[] $\theta_{t+1}\gets\theta_t$
            \end{enumerate*}
        \item increment $t$ and return to step~\ref{step: sample proposal}
        \end{enumerate}
    \end{enumerate}
\label{algorithm: Penalty Method Algorithm}
\end{algorithm}

\subsection{Prediction Calibration Using The Mini-Batch Size $n$}

As introduced in \sectionref{subsection: Expected posterior sampling with mini-batches}, the mini-batch size $n$ is a hyperparameter that we can use to calibrate the posterior predictive distribution. PBNN's ability to evaluate the likelihood over small mini-batches even in the presence of strong noise allows us to achieve this. A model is calibrated if, in the long-run, the proportion of forecast $\times$ percent credible intervals that succeed in covering the actual value of the predicted quantity turns out to be $\times$ percent \citep{10.2307/2987588}.

\begin{figure}[!ht]
    \centering
    \subfigure[Regression corresponding to\newline \figureref{fig:bnn_regression} and \figureref{fig:penalty_regression_batch_size_5}.]{
        \includegraphics[width=.45\textwidth]{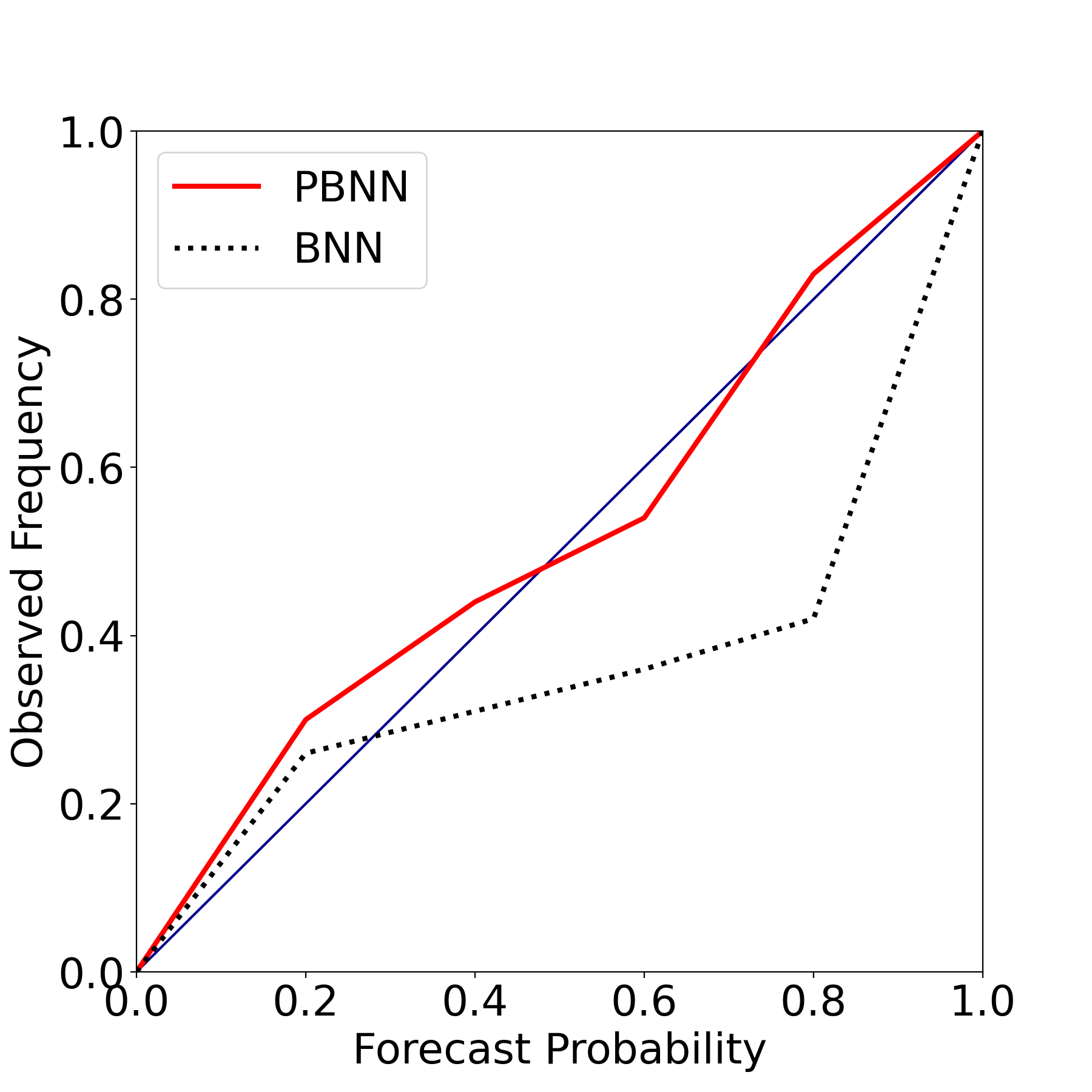}
    }
    \subfigure[MNIST classification.\newline $n=100$ and $M=10$]{
        \includegraphics[width=.45\textwidth]{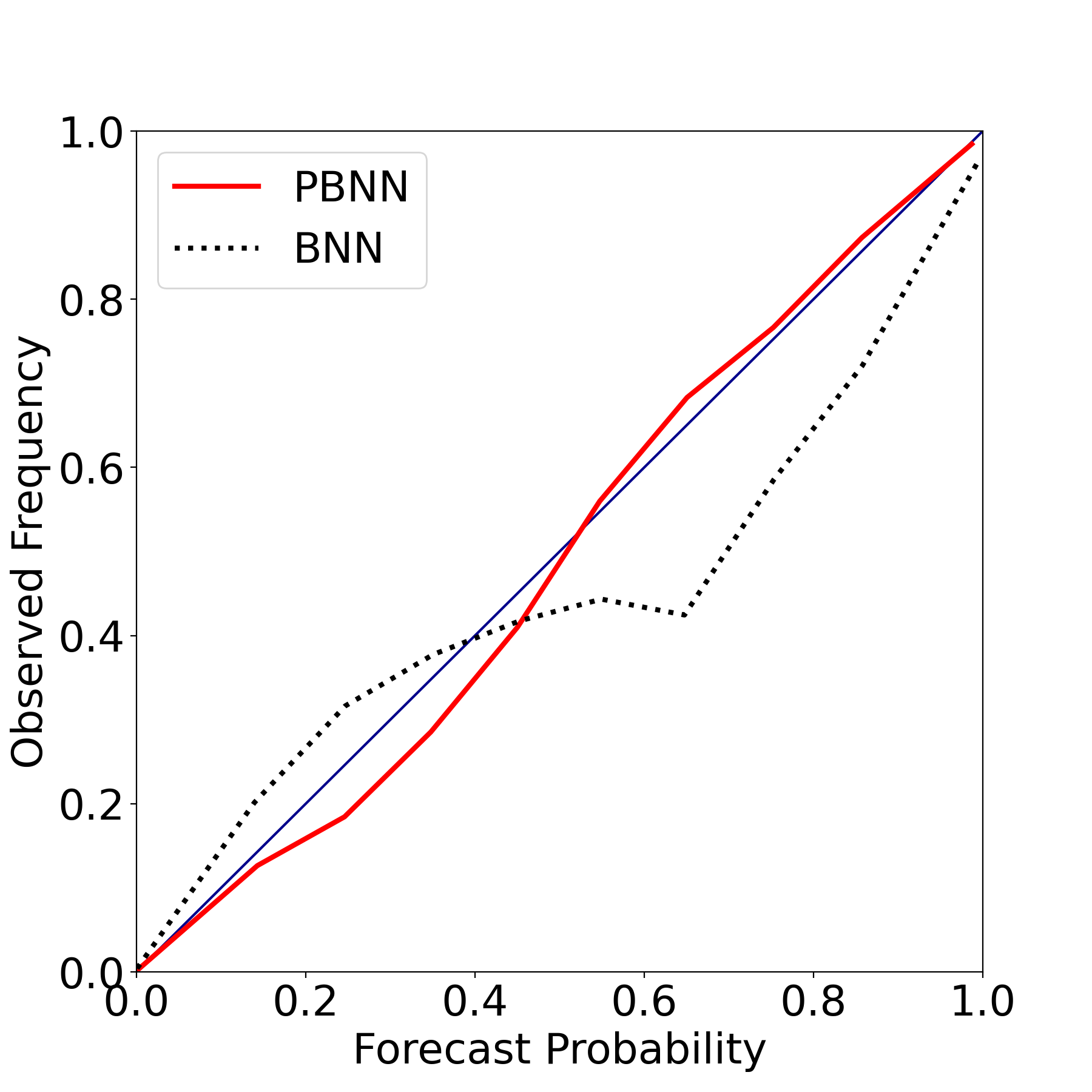}
        \label{figure: reliability_diagram_classification}
    }
    \caption{Reliability diagrams on test data. BNN prediction corresponds to \equationref{equation: predictive distribution} whereas the prediction of PBNN is computed using \equationref{equation: PBNN predictive posterior}. MNIST classifiers obtain a similar accuracy test score: 93.2\% using a PBNN, and 93.6\% using a BNN. The architecture of the softmax classifier is a single hidden layers containing 20 neurons.}
    \label{figure: reliability diagrams}
\end{figure}

The reliability diagrams in \figureref{figure: reliability diagrams} display the correspondence or discrepancy between a model prediction uncertainty and the observed frequencies in a test set of data. Ideal calibration corresponds to the $y=x$ diagonal: we see that PBNN mini-batch size can help calibrate a model prediction which is overconfident.

Note that, partial BNNs \citep{pmlr-v206-sharma23a} are used in the experiments shown in \figureref{figure: reliability diagrams} to induce overconfidence. Only the last layer is stochastic, such that the usual posterior distribution (\equationref{equation: posterior distribution}), given the full available dataset does not capture the total uncertainty (see misspecification in the \appendixref{apd:first}). In \figureref{figure: reliability diagrams}, BNN and PBNN share the same non-stochastic layers.

\section{PENALIZED LANGEVIN DYNAMICS}
\label{section: Penalized Langevin Dynamic}

\subsection{Metropolis Adjusted Langevin Algorithm}

In \algorithmref{algorithm: Penalty Method Algorithm}, we are free to choose a proposal distribution $q(\theta'|\theta)$. It is well known that choosing a non symmetric proposal distribution can speed up the mixing of the Markov Chain and help BNN scale to larger systems by maximizing the acceptance $A(\theta', \theta)$. Assuming a sufficiently small step from $\theta$ to $\theta'$, the Metropolis-Hastings acceptance writes:
\begin{equation}
    A(\theta', \theta)=\min\left(1,\frac{q(\theta|\theta')}{q(\theta'|\theta)}e^{-\Delta(\theta', \theta)}\right)\approx \min\left(1,\frac{q(\theta|\theta')}{q(\theta'|\theta)}e^{-(\theta-\theta')\cdot(\nabla_{\theta}\mathcal{L}_n(\theta')+\nabla_{\theta}\mathcal{L}_n(\theta))/2}\right)
\end{equation}
where we have Taylor-expanded the loss $\mathcal{L}(\theta)$ around $\theta'$. The maximization of $A(\theta', \theta)$ leads to a drift in the Gaussian proposal distribution that corresponds to the gradient of the loss:
\begin{equation}
    q(\theta'|\theta)=\mathcal{N}(\theta'; \theta-\eta\nabla_{\theta}\mathcal{L}_n(\theta), 2\eta)
    \label{equation: Langevin proposal}
\end{equation}
Sampling a new state $\theta'$ from the proposal distribution $q(\theta'|\theta)$ exactly corresponds to drawing a centered reduced normal random variable $\epsilon$, and computing:
\begin{equation}
    \theta'=\theta-\eta\nabla_{\theta}\mathcal{L}_n(\theta) + \sqrt{2\eta}\epsilon
    \label{equation: Langevin dynamics}
\end{equation}
\equationref{equation: Langevin dynamics} is equivalent to a Langevin dynamic and gives rises to the celebrated Metropolis-Adjusted Langevin algorithm (MALA) and Smart Monte Carlo \citep{10.1063/1.436415}.

\subsection{Penalized MALA}
In order to use a noisy gradient,  $\nabla_{\theta}\mathcal{L}_n(\theta)\simeq\nabla_{\theta}\mathcal{L}_n^n(\theta)$, as an approximation of the Gaussian mean's drift, one can design a proposal distribution $q(\theta|\theta')$ such that:
\begin{equation}
    q(\theta'|\theta)=\mathcal{N}(\theta'; \theta-\eta\nabla_{\theta}\mathcal{L}_n^n(\theta), 2\eta)
\end{equation}
and set a non zero step size $\eta$ while computing the Metropolis-Hastings acceptance:
\begin{equation}
    A(\theta', \theta)=\min\left(1,\frac{q(\theta|\theta')}{q(\theta'|\theta)}e^{-\delta({\theta',\theta})-\sigma^2({\theta',\theta})/2}\right)
    \label{equation: PBNN MH acceptance}
\end{equation}
which also satisfies the detailed balance in \equationref{equation: original detailed balance,equation: noisy detailed balance} on average. The non-trivial term in the Metropolis-Hastings acceptance then writes:
\begin{equation}
    \begin{split}
        \log\left(\frac{q(\theta|\theta')}{q(\theta'|\theta)}e^{-\delta({\theta',\theta})-\sigma^2({\theta',\theta})/2}\right)&=\frac{-1}{4\eta}\norm{\eta\left(\nabla_{\theta} \mathcal{L}_n^n(\theta')+\nabla_{\theta}\mathcal{L}_n^n(\theta)\right)-\sqrt{2\eta}\epsilon}^2\\
        &+\frac{1}{4\eta}\norm{\sqrt{2\eta}\epsilon}^2-\delta({\theta',\theta})-\sigma^2({\theta',\theta})/2
    \end{split}
\end{equation}

\subsection{Unadjusted Langevin Algorithm}

For large-sized models, biased samplers like the Unadjusted Langevin Algorithm (ULA) are known to be very effective. This is because they skip the rejection step i.e set $A(\theta', \theta)=1$ for a sufficiently small step size $\eta$
resulting in an unadjusted Langevin sampling as:
\begin{equation}
        \theta_{t+1}=\theta_{t}-\eta\nabla_\theta
        \mathcal{L}_n(\theta_t) + \sqrt{2\eta}\epsilon_t
        \label{equation: ULA}
\end{equation}
Note that in \equationref{equation: ULA}, new states are automatically accepted. Once again, in order to model a noisy estimate of the loss, it is tempting to replace the drift $\nabla_\theta\mathcal{L}_n(\theta_t)$  with an estimate such that:
\begin{equation}
    \eta\nabla_\theta\mathcal{L}_n(\theta)=\eta\nabla_\theta \mathcal{L}_n^n(\theta)+\eta\sigma(\theta)
\end{equation} 
For a vanishing step size $\eta\to 0$, one may then expect that the additional noise term, $\eta \sigma(\theta)$, gets
negligible compared to the random noise of order $\eta^{1/2}$ in \equationref{equation: ULA} as suggested by SGLD \citep{10.5555/3104482.3104568}.
However, the uncertainty of the loss gradient $\sigma(\theta)$ does in general not result in white noise, but is correlated between different parameters $\theta$. For non-vanishing $\eta$, the noisy loss gradient can thus trigger a significant departure from the target posterior (see \citet{brosse_promises_2018} and \figureref{fig:sgld_1e-7.png,fig:sgld_1e-8.png} in \appendixref{apd:first}).

\section{CONCLUSION}

Uncertainty quantification for the predictions of large size neural networks remains an open issue. In this work, we have introduced PBNN as a novel approach to enable data subsampling for Bayesian Neural Network without relying on a noisy estimate of the gradient such as those used in Stochastic Gradient Langevin Dynamic.

First, we demonstrated that a naive estimation of the likelihood based on noisy loss introduces a bias in the posterior sampling if not correctly accounted for. We then showed that a generalization of the Metropolis Hastings algorithm can eliminate the bias, allowing for exact posterior sampling even amid substantial noise. This approach necessitates an additional "noise penalty" that matches the variance of the noisy loss difference. The drawback of the method is that it exponentially suppresses the MCMC acceptance probability.

In practice, the noise penalty corresponds to replacing a single large data set by multiple smaller subsampled mini-batches associated with an uncertainty over their losses. Varying the size of the mini-batches enables a calibration of the predictive model. We have compared this benefit of PBNN to other techniques such as tempered Safe Bayes approaches.

Based on this calibration principle, we conducted numerical experiments demonstrating the robust predictive performance of PBNNs. We are optimistic that integrating data subsampling with other Monte Carlo acceleration techniques such as HMC \citep{neal_mcmc_2010} will enable the computation of uncertainties for previously unattainable models and data set sizes.

Lastly, PBNN appears to be particularly well-suited for scenarios where the data sets $\mathcal{D}$ are distributed across multiple decentralized devices as in the typical federated learning setup. The key to its suitability lies in the noise penalty which is determined by the variance of losses across several mini-batches. These losses can be computed independently on each device's data set. Consequently, PBNN should enable the possibility to compute uncertainty with separate data sets without exchanging them.

%
%
%

\section{ACKNOWLEDGMENTS}
This work has been supported by the French government under the "France 2030” program, as part of the SystemX Technological Research Institute.
Authors thank Victor Berger and Han Wang for useful comments and discussions.

\bibliography{refs}

\newpage

\appendix

\section{SGLD, Safe Bayes and PBNN}
\label{apd:first}

\begin{figure}[!ht]
    \centering
    \subfigure[SGLD prediction,\newline
    mini-batch size $n=5$\newline
    $\eta=10^{-7}$.]{
        \includegraphics[width=.45\textwidth]{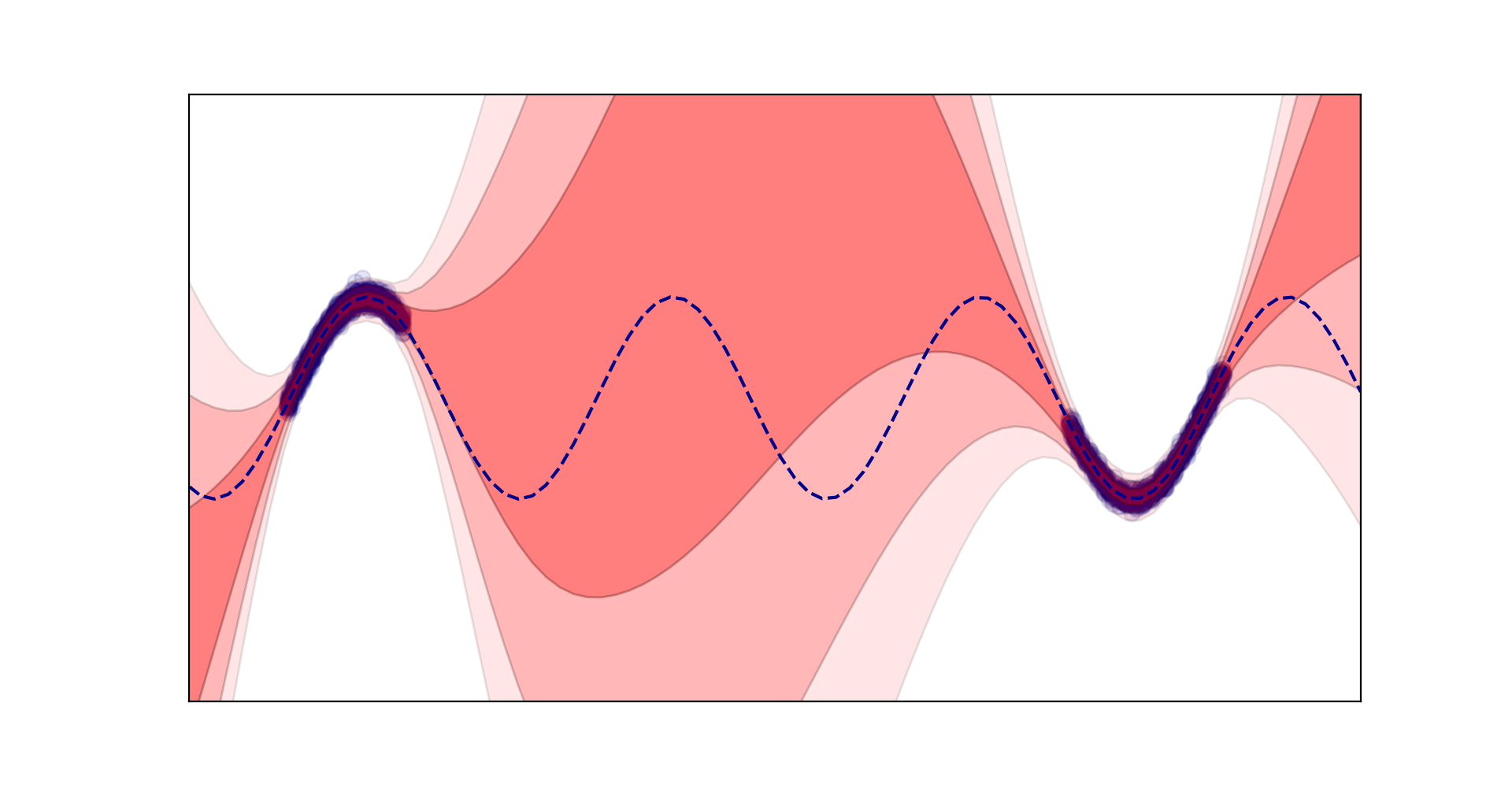}
        \label{fig:sgld_1e-7.png}
    }
    \subfigure[SGLD prediction\newline
    mini-batch size $n=5$\newline
    $\eta=10^{-8}$.]{
        \includegraphics[width=.45\textwidth]{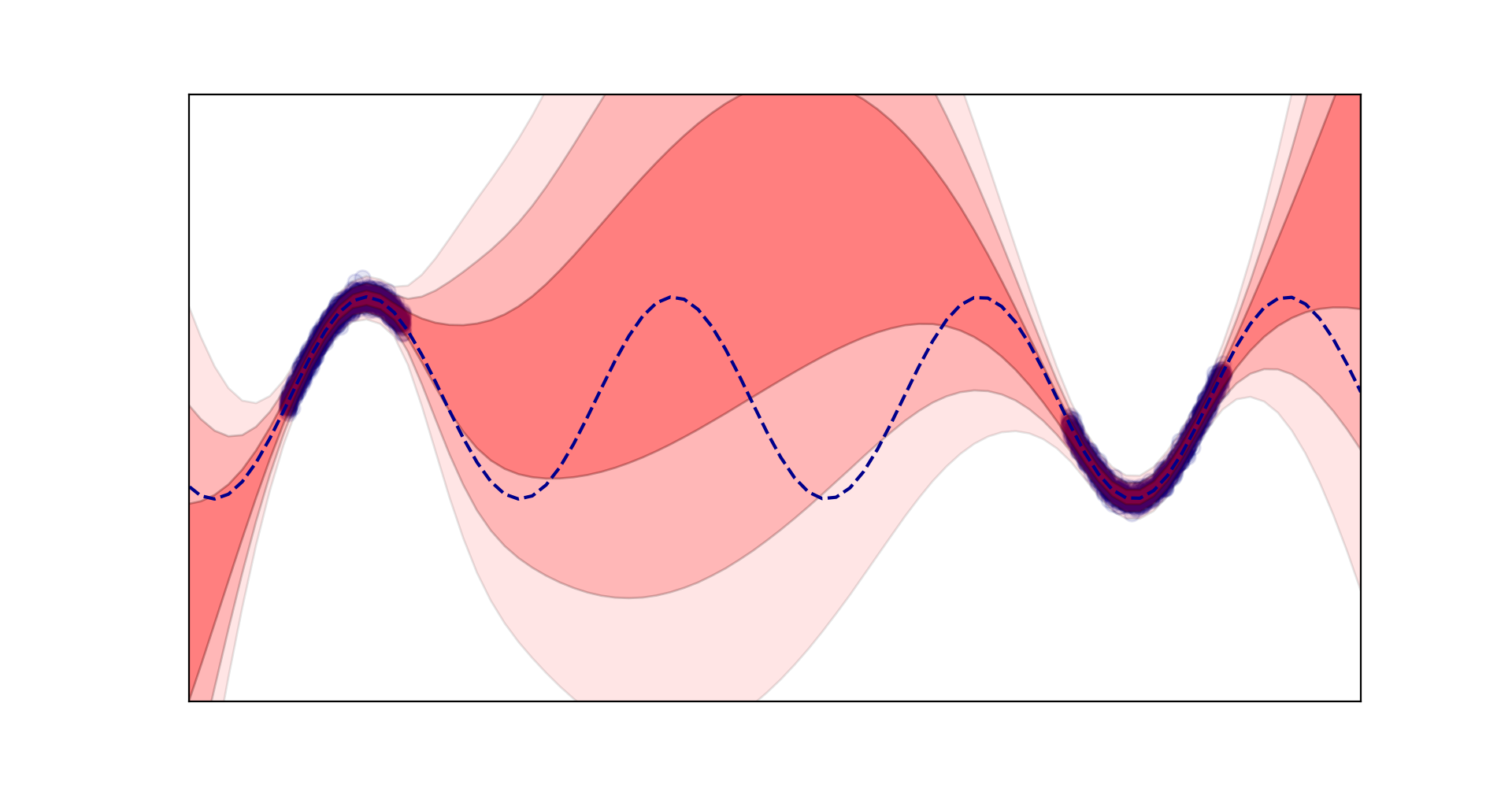}
        \label{fig:sgld_1e-8.png}
    }
    \subfigure[Tempered BNN with $n=5$]{
        \includegraphics[width=.45\textwidth]{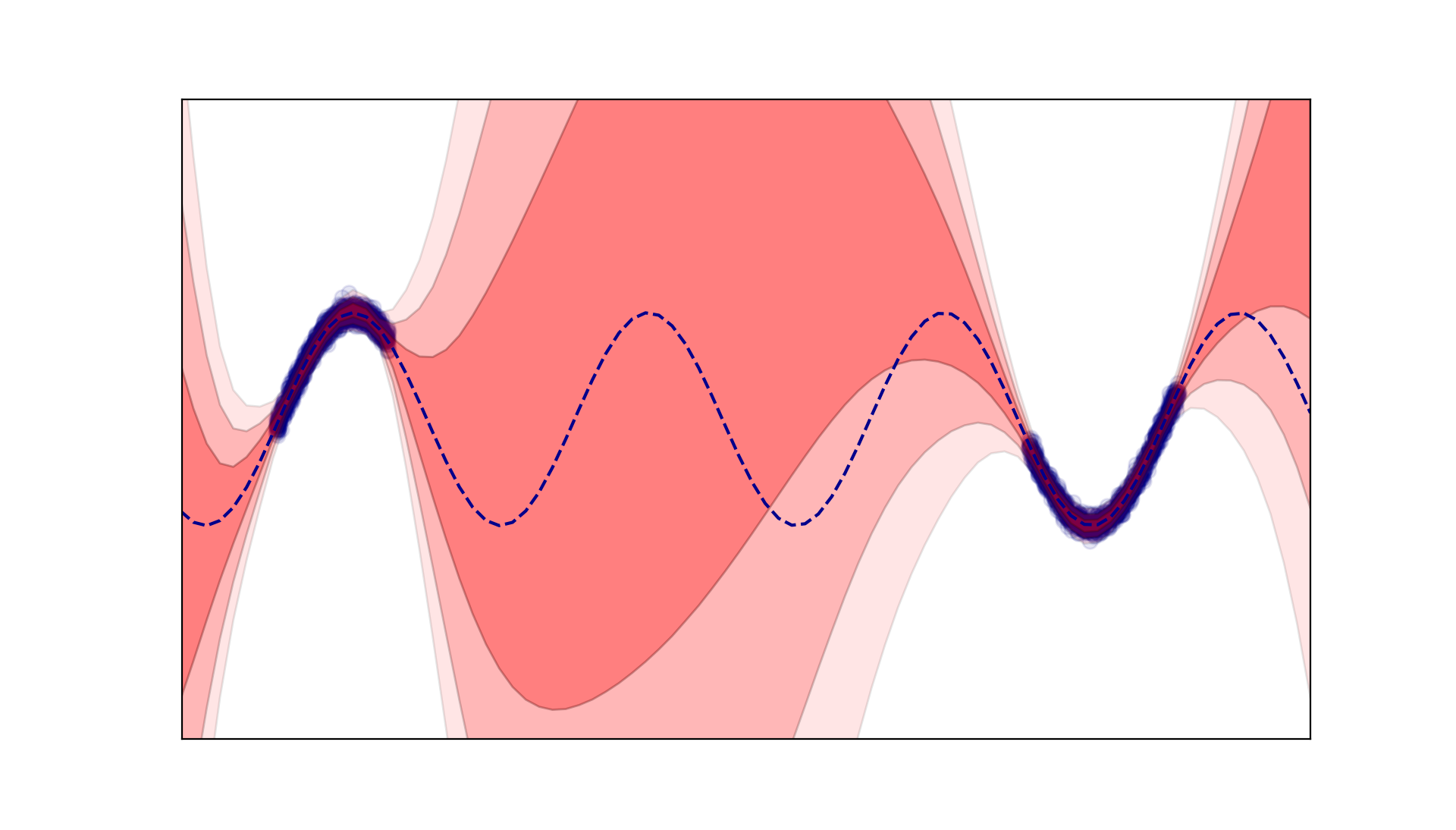}
        \label{figure: tempered}
    }
    \subfigure[PBNN prediction,\newline
    mini-batch size $n=5$.]{
        \includegraphics[width=.45\textwidth]{figures/penalty_regression_batch_size_5.png}
        \label{figure: PBNN reference}
    }
    \caption{SGLD results are very sensitive to the learning rate $\eta$. \figureref{figure: PBNN reference} is used as a reference as it targets the same posterior as \figureref{fig:sgld_1e-7.png} and \figureref{fig:sgld_1e-8.png}. We note that as expected, Safe Bayes approaches obtain a qualitatively different result from PBNN for the same mini-batch size.}
    \label{figure: SGLD}
\end{figure}

In principle SGLD can be used to compute \equationref{equation: PBNN predictive posterior} in the same way as PBNN. However, because of the noise introduced by the data subsampling, SGLD requires a vanishing learning rate $\eta$ as shown in \figureref{figure: SGLD}. In the same figure, PBNN has a learning rate of $\eta=0.1$: varying $\eta$ changes the MCMC acceptance and not the prediction which is independent of the step size.

\end{document}